\documentstyle[aclap,psfig]{article}
\pssilent

\title{A Complexity Measure for Diachronic Chinese Phonology}
\author{
  Anand Raman \\
  Computer Science \\
  Massey University\\
  Palmerston North\\New Zealand\\
  {\tt A.Raman@massey.ac.nz}\And
  John Newman \\
  Linguistics and SLT\\
  Massey University\\
  Palmerston North \\New Zealand\\
  {\tt J.Newman@massey.ac.nz}\And
  Jon Patrick\\
  Information Systems\\
  Massey University\\
  Palmerston North \\New Zealand\\
  {\tt J.D.Patrick@massey.ac.nz}
}
\begin{document}
\newcommand{\zihui} {{\em Zihui}}
\newcommand{\qie}{{\em Qie yun}}
\maketitle
\begin{abstract}
This paper addresses the problem of deriving distance
measures between parent and daughter languages with specific relevance
to historical Chinese phonology.  The diachronic relationship between
the languages is modelled as a Probabilistic Finite State Automaton.
The Minimum Message Length principle is then employed to find the
complexity of this structure.  The idea is that this measure is
representative of the amount of dissimilarity between the two
languages.
\end{abstract}

\section{Introduction}

When drawing up genetic trees of languages, it is sometimes useful to
quantify the degree of relationship between them.  Mathematical
approaches along these lines have been pursued for some time now ---
\newcite{Embleton:MMG91} is an excellent review of some important
techniques.  \newcite{Cheng:QCD82}, in fact, attempts to address the
issue central to this paper --- that of obtaining distance measures
between related Chinese dialects.  However, he does this at a lexical
level by using Karl Pearson's tetrachoric correlation coefficient on
905 words from a lexical dictionary \cite{cihui64}.  This paper takes
a novel approach to this problem by pioneering the use of phonological
data to find dissimilarity measures, as opposed to lexical (which has
been used most frequently up till now), semantic or syntactic
data.\footnote{Indeed, semantic similarity, which is usually necessary
for the identification of cognates in Indo-European languages, is not
even relevant in the case of the Chinese languages we are concerned
with in this paper because cognates can be visually identified in
Chinese languages due to a common ideographic writing system
stretching back over 3 millenia \cite[p.103]{Streeter:DOC77}.}

An argument can also be made that phonetic or phonological
dissimilarity measures, being the least abstract of all, could give
the most realistic results.  Unfortunately, studies in this direction
have been relatively rare.  Two such works which should be mentioned
are \newcite{Grimes:LDR1959} and \newcite{Hsieh:NMD73}, both of which
are, however, constrained by the use of lexicostatistical methodology.
In fairness to existing methods, it must be noted that many other
existing methods for obtaining dissimilarity measures are in fact
applicable to non-lexical data for deriving non-lexical measures.  In
practice, though, they have been constrained by a preoccupation with
the lexicon as well as by the unavailability of phonological
data.\footnote{This was also pointed out by Professor Sheila Embleton,
York University, Toronto in a personal communication: Comment on using
a phonological dissimilarity measure. In email correspondence dt. 9
Oct 1994.}  Hopefully, the phonological data developed in this project
should provide fresh input to those methods and revive their
application to the problem area in future research.

\section{Data}

The data we use to illustrate our ideas are two phonological histories
taken from the field of Chinese linguistics.  One is an account of the
Modern Beijing (MB) dialect from an earlier stage of Chinese, referred
to as Middle Chinese, and published as \newcite{Chen:MCMP76}; the
other is an account of the Modern Cantonese (MC) dialect also from
Middle Chinese, published as Chen and Newman (1984a, 1984b and 1985).
\nocite{Chen:MCMC-I84,Chen:MCMC-II84,Chen:MCMC-III85} These should be
consulted for further explanation of the diachronic rules and their
relative chronology as well as for an explanation of the rule labels
used in this paper.  For brevity, we will refer to the former as
Chen76 and the latter as CN84 in subsequent sections.  We would now
like to draw attention to five features of these accounts which make
them ideal for the purpose at hand:

\begin{enumerate}

\item The accounts are relatively explicit in their expositions. Each
account assumes Middle Chinese reconstructions which are phonetically
explicit, states each rule in a formal style, and defines the ordering
relationships which hold between the rules. This degree of
comprehensiveness and explicitness in writing the history of a
language is relatively rare. It is even rarer to have accounts of two
related dialects described in a similarly explicit way.  Obviously,
when it comes to translating historical accounts into phonological
derivations, the more explicit the original account, the more readily
one can arrive at the derivations.

\item The two accounts assume identical reconstructions for the Middle
Chinese forms, which of course is crucial in any meaningful comparison
of the two dialects. Not surprisingly, given the existence of Sinology
as an established field and one with a history going back well over a
hundred years, there are many conflicting proposals about Middle
Chinese and its pronunciation. Decisions about the forms of Middle
Chinese go hand in hand, necessarily, with corresponding decisions
about the historical rules which lead from those forms to modern-day
reflexes. One can not easily compare competing historical accounts if
they assume different reconstructed forms as their starting points.
See Chen76 for a full description and justification of the Middle
Chinese reconstructions used in these accounts.

\item The two accounts are couched in terms of one phonological
framework. This, too, is a highly desirable feature when it comes to
making comparisons between the sets of rules involved in each
account. The framework could be described as a somehwat ``relaxed''
version of SPE \cite{Chomsky:SPE68}. For example, the accounts make
use of orthodox SPE features alongside others where it was thought
appropriate (e.g. [+/- labial], [+/- acute]). Phonotactic conditions
are utilized as a way of triggering certain phonological changes,
alongside more conventional rule statements.

\item The accounts purport to describe the phonological histories of a
single database of Chinese characters and their readings in modern
dialects \cite{zihui62}.  This is a substantial database containing
about 2,700 Chinese characters and it is the readings of these
characters in two of the dialects --- Modern Beijing and Modern
Cantonese which are the outputs of the rule derivations in the two
accounts.

\item The accounts themselves are published in an easily available
journal, {\em The Journal of Chinese Linguistics}, which allows
readers to scrutinize the original discussion and rule statements.

\end{enumerate}

The features alluded to in points 1--5 make these two accounts
uniquely suited to testing out formal hypotheses relating to
historical phonology.  The historical account of Modern Beijing/Modern
Cantonese is construed as a set of derivations. The input to a
derivation is a reconstructed Middle Chinese form; the input is
subjected to a battery of (ordered) phonological rules; and the output
of the derivation is the reflex in the modern dialect.

\section{Modelling Phonological Complexity}
\label{sec:mod}

The mechanistic model we have used to represent diachronic
phonological derivations is that of Probabilistic Finite State
Automata (PFSA).  These are state determined machines which have
stochastic transition functions.  The derivation of each word in MB or
MC from Middle Chinese consists of a sequence of diachronic rules.
These rule sequences for each of the approximately 2700 words are used
to construct our PFSA.  Node 0 of the PFSA corresponds to the
reconstructed form of the word in Middle Chinese.  Arcs leading out of
states in the PFSA represent particular rules that were applied to a
form at that state, transforming it into a new intermediate form.  A
transition on a delimiter symbol, which always returns to state 0,
signifies the end of a derivation process whereby the final form in
the daughter language has been arrived at.  The weightings on the arcs
represent the number of times that particular arc was traversed in
processing the entire corpus of words.  The complete PFSA then
represents the phonological complexity of the derivation process from
Middle Chinese into one of the modern dialects.

If this is the case, then the length of the minimal description of the
PFSA would be indicative of the distance between the parent and
daughter languages.  There are two levels at which the diachronic
complexity can be measured.  The first is of the canonical PFSA, which
is a trie encoding of the rules.  This is the length of the diachronic
phonological hypothesis accounting for the given dataset.  The second
is of a minimised version of the canonical machine.  Our minimisation
is performed initially using the sk-strings method of
\newcite{Raman:SKS97} and then reducing the resultant automaton
further with a beam search heuristic \shortcite{Raman:BSS97}.  The
sk-strings method constructs a non-deterministic finite state
automaton from its canonical version by successively merging states
that are indistinguishable for the top s\% of their most probable
output strings limited to a length of $k$ symbols.  Both {\em s\/} and
{\em k\/} are variable parameters that can be set when starting
program execution.  In this paper, the reduced automata are the best
ones that could be inferred using any value of string size ($k$) from
1 to 10 and any value of the agreement percentage ($s$) from 1 to 100.
The beam search method reduces the PFSA by searching recursively
through the best $m$ descendants of the current PFSA where a
descendant is defined to be the result of merging any two nodes in the
parent PFSA.  The variable parameter $m$ is called the beam size and
determines the exhaustiveness of the search.  In this paper, $m$ was
set to 200, which was the maximum the Sun Sparcserver 1000 with 256 MB
of main memory could tolerate.

The final resultant PFSA, minimised thus is, strictly speaking, a
generalisation of the proposed phonology.  Its size is not really
indicative of the complexity of the original hypothesis, but it serves
to bring to light important patterns which repeat themselves in the
data.  The minimisation, in effect, forms additional diachronic rules
and highlights regular patterns to a linguist.  The size of this
structure is also given in our results to show the effect of further
generalisation to the linguistic hypothesis.

A final point needs to be made regarding the motivation for the
additional sophistication embodied in this method as compared to, say,
a more simplistic phonological approach like a distance measure based
on a simple summation of the number of proposed rules.  Our method not
only gives a measure dependent on the number of rules, but also on the
inter-relationship between them, or the regularity present in the
whole phonology.  A lower value indicates the presence of greater
regularity in the derivation process.  As a case in point, we may look
at two closely related dialects, which have the same number of rules
in their phonology from a common parent.  It may be the case that one
has diverged more by losing more of its original structure.  As in the
method of internal reconstruction, if we assume that the complexity of
a language increases with time due to the presence of residual forms
\cite[p.150--153]{Crowley:IHL87}, the PFSA derived for the more
distant language will have a greater complexity than the other.

\section{Procedural Decisions}

The derivations that were used in constructing the PFSA were traced
out individually for each of the 2714 forms and entered into a
spreadsheet for further processing.  The Relative Chronologies (RC) of
the diachronic rules given in Chen76 and CN84 propose rule orderings
based on bleeding and feeding relationships between rules.\footnote{If
rule A precludes rule B from applying by virtue of applying before it,
then A is said to bleed B.  If rule A causes rule B to apply by
applying before it, it is said to feed rule B.}  We have tried to be
as consistent as possible to the RC proposed in Chen76 and CN84.  For
the most part, we view violations to the RC as exceptions to their
hypothesis.  Consistency with the RC proposed in Chen76 and CN84
has been maintained as far as possible.  For the most part, violations
to them are viewed as serious exceptions.  Thus if Rule A is ordered
before Rule B in the RC, but is required to apply after Rule B in a
specific instance under consideration, it is made an exceptional
application of Rule A, denoted by ``[A]''.  Such exceptional rules are
considered distinct from their normal forms.  The sequence of rules
deriving Beijing {\em tou\/} from Middle Chinese {\em *to\/}
(``all''), for example, is given as
``t1-split:raise-u:diphthong-u:chamel:''.  However, ``diphthong-u'' is
ordered before ``raise-u'' in the RC.  The earlier rule in the RC is
thus made an exceptional application and the rule sequence is given
instead as ``t1-split:raise-u:[diphthong-u]:chamel:''.

There are also some exceptional phonological changes not accounted for
by CN84 or Chen76.  In these cases, we form a new rule representing
the change that took place, denote it in square brackets to show its
exceptional status.  Related exceptions are grouped together as a
single exceptional rule.  For example, Tone-4 in Middle Chinese only
changes to Tone-1a or Tone-2 in Beijing when the form has a voiceless
initial.  However, for the Middle Chinese form {\em *niat\/} (``pinch
with fingers'') in Tone-4, the corresponding Beijing form is {\em
nie\/} in Tone-1a.  Since the n-initial is voiced, the t4-tripart rule
is considered to apply exceptionally.  The complete rule sequence is
thus denoted by ``raise-i:apocope:chamel:[t4]:'' where the ``[t4]''
exceptional rule covers cases when Tone-4 in SMC unexpectedly changed
into Tone-1a or Tone-2 in Beijing in the absence of a voiceless
initial.

It also needs to be mentioned that there are a few cases where an
environment for the application of a rule might exist, but the rule
itself may not apply although it is required to by the linguistic
hypothesis.  This would constitute an exception again.  The details of
how to handle this situation more accurately are left as a topic for
future work, but we try to account for it here by applying a special
rule [!A] where the `!' is meant to indicate that the rule A didn't
apply when it ought to have.  As an example, we may consider the
derivation of Modern Cantonese {\em hap\/}(Tone 4a) from Middle
Chinese {\em *k\textsuperscript{h}ap}(Tone 4) (``exactly'').  The
sequence of rules deriving the MC form is
``t4-split:spirant:x-weak:''.  However, since the environment is
appropriate (voiceless initial) for the application of a further rule,
AC-split, after t4-split had applied, the non-application of this
additional rule is specified as an exception.  Thus,
``t4-split:spirant:x-weak:[!AC-split]:'' is the actual rule sequence
used.

In general, the following conventions in representing and treating
exceptions have been followed as far as possible: Exceptional rules
are always denoted in square brackets.  They are considered excluded
from the RC and thus are consistently ordered at the end of the rest
of the derivation process wherever possible.

A final detail concerns the status of allophonic changes in the
phonology.  The derivation process is actually two-stage, comprising a
diachronic phase during which phonological changes take place and a
synchronic phase during which allophonic changes are automatically
applied.  Changes caused by Cantonese or Beijing Phonotactic
Constraints (PCs) are treated as allophonic rules and fall into the
synchronic category, whereas PCs applying to earlier forms are treated
in line with the regular diachronic rules which Chen76 calls P-rules.

A minor problem presents itself when it comes to making a clear-cut
separation between the historical rules proper and the synchronic
allophonic rules.  In Chen76 and CN84, they are not really considered
part of the historical derivation process.  Yet it was found that the
environment for the application of a diachronic rule is sometimes
produced by an allophonic rule.  Such feeding relationships between
allophonic and diachronic rules make the classification of those
allophonic rules difficult.

The only rule considered allophonic in Beijing is the *CHAMEL PC, this
being a rule which determines the exact qualities of MB vowels.  For
Cantonese, CN84 has included two allophonic rules within its RC under
bleeding and feeding relationships with P-rules.  These are the
BREAK-C and Y-FUSE rules, both of which concern vocalic detail.  In
these cases, every instance of their application within the diachronic
phonology has been treated as an exception, effectively elevating
these exceptions to the status of diachronic rules.  In other cases,
as with other allophonic rules, they are always ordered after all the
diachronic rules.  Since the problem regarding the status of
allophonic rules in general is properly in the domain of historical
linguists, it is beyond the scope of this work.  It was thus decided
to provide two complexity measures --- one including allophonic detail
and one excluding all allophonic detail not required for the
derivation process.

\section{Minimum Message Length}

The Minimum Message Length (MML) principle of \newcite{Georgeff:GSC84}
is used to compute the complexity of the PFSA.  For brevity, we will
henceforth call the Minimum Message Length of PFSA as the MML of PFSA
or where the context serves to disambiguate, simply MML.

In the context of data transmission, the MML of a set of symbols is
the minimum number of bits needed to transmit a static model together
with the data symbols given this model {\em a priori}.  In the context
of PFSA, the MML is a sum of:
\begin{itemize}
\item  the length of encoding a description of the proposed machine
\item  the length of encoding the dataset assuming it was emitted by the
       proposed machine
\end{itemize}
The following formula is used for the purpose of computing the MML:
\begin{eqnarray*}
\sum_{j=1}^{N}\{ m_j + 
	\log \frac{(t_j-1)!}{(m_j-1)!\prod\limits_{i=1}^{m_j}(n_{ij}-1)!} + \\
		m_j \log V + m_j'\log N \} - \log (N-1)!
\end{eqnarray*}
\noindent where $N$ is the number of states in the PFSA, $t_j$ is the
number of times the $jth$ state is visited, $V$ is the cardinality of
the alphabet including the delimiter symbol, $n_{ij}$ the frequency of
the $ith$ arc from the $jth$ state, $m_j$ is the number of different
arcs from the $jth$ state and $m_j'$ is the number of different arcs
on non-delimiter symbols from the $jth$ state.  The logs are to the
base 2 and the MML is in bits.

The MML formula given above assumes a non-uniform prior on the
distribution of outgoing arcs from a given state.  This contrasts with
the MDL criterion due to \newcite{Rissanen:MSD78} which recommends the
usage of uniform priors.  The specific prior used in the specification
of $m_j$ is $2^{-m_j}$, i.e. the probability that a state has $n$
outgoing arcs is $2^{-n}$.  Thus $m_j$ is directly specified in the
formula using just $m_j$ bits and the rest of the structure
specification assumes this.  It is also assumed that targets of
transitions on delimiter symbols return to the start state (State 0
for example) and thus don't have to be specified.  The formula is a
modification for non-deterministic automata of the formula in
\newcite{Patrick:RII87} where it is stated with two typographical errors
(the factorials in the numerators are absent).  It is itself a
correction (through personal communication) of the formula in
\newcite{Wallace:GOI84} which follows on from work in numerical taxonomy
\cite{Wallace:IMC68} that applied the MML principle to derive
information measures for classification.

\section{Results}

The results of our analysis are given in Tables~\ref{tbl:canon-mml}
(for canonical PFSA) and~\ref{tbl:reduced-mml} (for reduced PFSA).
Row 1 represents PFSA which have only diachronic detail in them and
Row 2 represents PFSA which do not distinguish between diachronic and
allophonic detail.  Column 1 represents the MML of the PFSA derived
for Modern Cantonese and and column 2 represents the MML of PFSA for
Modern Beijing.  As mentioned in Section~\ref{sec:mod}, smaller values
of the MML reflect a greater regularity in the structure.

\begin{table}[htb]
\begin{center}
\begin{tabular}{|l|p{2cm}|p{2cm}|} \hline
		& {\bf Cantonese}	   & {\bf Beijing}  \\ \hline\hline
Diachronic	& 35243.58 bits		   & 36790.93 bits            \\
only		& (1168 states, 1167 arcs) & (1212 states, 1211 arcs) \\ \hline
Diachronic +	& 37782.43 bits		   & 39535.43 bits  	      \\
Allophonic	& (1321 states, 1320 arcs) & (1468 states, 1467 arcs) \\ \hline
\end{tabular}
\end{center}
\caption{MMLs for the canonical PFSA for Middle Chinese to Modern
   Cantonese and Modern Beijing respectively }
\label{tbl:canon-mml}
\end{table}

\begin{table}[htb]
\begin{center}
\begin{tabular}{|l|p{2cm}|p{2cm}|} \hline
		& {\bf Cantonese}	 & {\bf Beijing}  \\ \hline\hline
Diachronic	& 30379.01 bits		 & 30366.55 bits  	  \\
only		& (174 states, 640 arcs) & (142 states, 595 arcs) \\ \hline
Diachronic +	& 32711.49 bits		 & 31585.79 bits  	  \\
Allophonic	& (195 states, 707 arcs) & (153 states, 634 arcs) \\ \hline
\end{tabular}
\end{center}
\caption{MMLs for the reduced PFSA for Middle Chinese to Modern 
         Cantonese and Modern Beijing respectively }
\label{tbl:reduced-mml}
\end{table}

The canonical PFSA are too large and complex to be printed on A4 paper
using viewable type.  However, it is possible to trim off some of the
low frequency arcs from the reduced PFSA to alleviate the problem of
presenting them graphically.  Thus the reduced PFSA for Modern Beijing
and Modern Cantonese are presented in Figures~\ref{fig:Mand-opfsa}
and~\ref{fig:Cant-opfsa} at the end of this paper, but arcs with a
frequency less than 10 have been pruned from them.  Since several arcs
have been pruned, the PFSA may not make complete sense as some nodes
may have outgoing transitions without incoming ones and vice-versa.
There is further a small amount of overprinting.  They are solely for
the purposes of visualisation of the end-results and not meant to
serve any other useful purpose.  The arc frequencies are indicated in
superscript font above the symbol, except when there is more than one
symbol on an arc, in which case the frequencies are denoted by the
superscript marker ``\^{ }''.  Exclamation marks (``!'') indicate arcs
on delimiter symbols to state 0 from the state they superscript.
Their superscripts represent the frequency.

Superficially, the PFSA may seem to resemble the graphical
representation of the Relative Chronologies in Chen76 and CN84, but in
fact they are more significant.  They represent the actual sequences
of rules used in deriving the forms rather than just the ordering
relation among them.  The frequencies on the arcs also give an idea of
how many times a particular rule was applied to a word at a certain
stage of its derivation process.  Certain rules that rarely apply may
not show up in the diagram, but that is because arcs representing them
have been pruned.  The MML computation process, however, accounted for
those as well.

The complete data corpus, an explanation of the various exceptions to
rules and the programs for constructing and reducing PFSA are
available from the authors.

\section{Discussion}

The results obtained from the MMLs of canonical machines show that
there is a greater complexity in the diachronic phonology of Modern
Beijing than there is in Modern Cantonese.  These complexity measures
may be construed as measures of distances between the languages and
their ancestor.  Nevertheless we exercise caution in interpreting the
results as such.  The measures were obtained using just one of many
reconstructions of Middle Chinese and one of many proposed diachronic
phonologies.  It is, of course, hypothetically possible that a
simplistic reconstruction and an overly generalised phonology could
give smaller complexity measures by resulting in less complex PFSA.
One might argue that this wrongly indicates that the method of
obtaining distances as described here points to the simplistic
reconstruction as the better one.  This problem arises partly because
of the fact that the methodology outlined here assumes all linguistic
hypotheses to be equally likely {\em a-priori}.  We note, however,
that simplicity and descriptive economy are not the only grounds for
preferring one linguistic hypothesis to another
\cite[p.47]{Bynon:HL83}.  Many other factors are usually taken into
consideration to ensure whether a reconstruction is linguistically
viable.  Plausibility and elegance \cite[p.314]{Harms:SRD90},
knowledge of what kinds of linguistic changes are likely and what are
unlikely \cite[p.90]{Crowley:IHL87}, and in the case of Chinese,
insights of the ``Chinese philological tradition'' \cite{Newman:ECP87}
are all used when deciding the viability of a linguistic
reconstruction.  Thus, a final conclusion about the linguistic problem
of subgrouping is still properly within the domain of historical
linguists.  This paper just provides a valuable tool to help quantify
one of the important parameters that is used in their decision
procedure.

We make a further observation about the results that the complexity
measures for the phonologies of Modern Beijing and Modern Cantonese
are not immensely different from each other.  Interestingly also,
while the MML of the canonical PFSA for Modern Beijing is greater than
that for Modern Cantonese, the MML of the reduced PFSA for Modern
Beijing is less than that for Modern Cantonese.  While the differences
might be within the margin of error in constructing the derivations
and the PFSA, it is possible to speculate that the generalisation
process has been able to discern more structure in the diachronic
phonology of Modern Beijing than in Modern Cantonese.  From a
computational point of view, one could say that the scope for further
generalisation of the diachronic rules is greater for Modern Cantonese
than for Modern Beijing or that there is greater structure in the
evolution of Modern Beijing from Middle Chinese than in the evolution
of Cantonese.  One could perhaps claim that this is due to the extra
liberty taken historically by current Modern Cantonese speakers to
introduce changes into their language as compared to their Mandarin
speaking neighbours.  But it would be na\"{\i}ve to conclude so here.
The study of the actual socio-cultural factors which would have
resulted in this situation is beyond the scope of this paper.

It is also no surprise that the MMLs obtained for the two languages
are not very different from each other although the difference is
large enough to be statistically significant.\footnote{We are grateful
to an anonymous reviewer for raising the question of what the smallest
difference in MML would be before having significance.  At least one
of the present authors claims the difference in MML for a single
set of data to be approximately an odds ratio.  Thus, a difference of
$n$ bits (however small $n$ is) would point to an odds ratio of
1:$2^n$ that the larger PFSA is more complex than the smaller one.
The explanation is not directly applicable in this case as we are
comparing two different data sets and so further theoretical
developments are necessary.}  Indeed, this is to be expected as they
are both contemporary and have descended from a common ancestor.  We
can expect more interesting results when deriving complexity measures
for the phonologies of languages that are more widely separated in
time and space.  It is here that the method described in this paper
can provide an effective tool for subgrouping.

\section{Conclusion and Future Work}

In this paper, we have provided an objective framework which will
enable us to obtain distance measures between related languages.  The
method has been illustrated and the first step towards actually
applying it for historical Chinese linguistics has also been taken.
It has been pointed out to us, though, that the methodology described
in this paper could in fact be put to better use than in just deriving
distance measures.  The suggestion was that it should be possible, in
principle, to use the method to choose between competing
reconstructions of protolanguages as this tends to be a relatively
more contentious area than subgrouping.

It is indeed possible to use the method to do this --- we could retain
the basic procedure, but shift the focus from studying two descendants
of a common parent to studying two proposed parents of a common set of
descendants.  A protolanguage is usually postulated in conjunction
with a set of diachronic rules that derive forms in the descendant
languages.  We could thus use the methodology described in this
paper to derive a large number of forms in the descendant languages
from each of the two competing protolanguages.  Since descriptive
economy is one of the deciding factors in selecting historical
linguistic hypotheses, the size of each body of derivations, suitably
encoded in the form of automata, in conjunction with other linguistic
considerations will then give the plausibility of that reconstruction.
Further study of this line of approach is, however, left as a topic for
future research.

\begin{onecolumn}
\begin{figure}[htb]
  \centerline{\psfig{figure=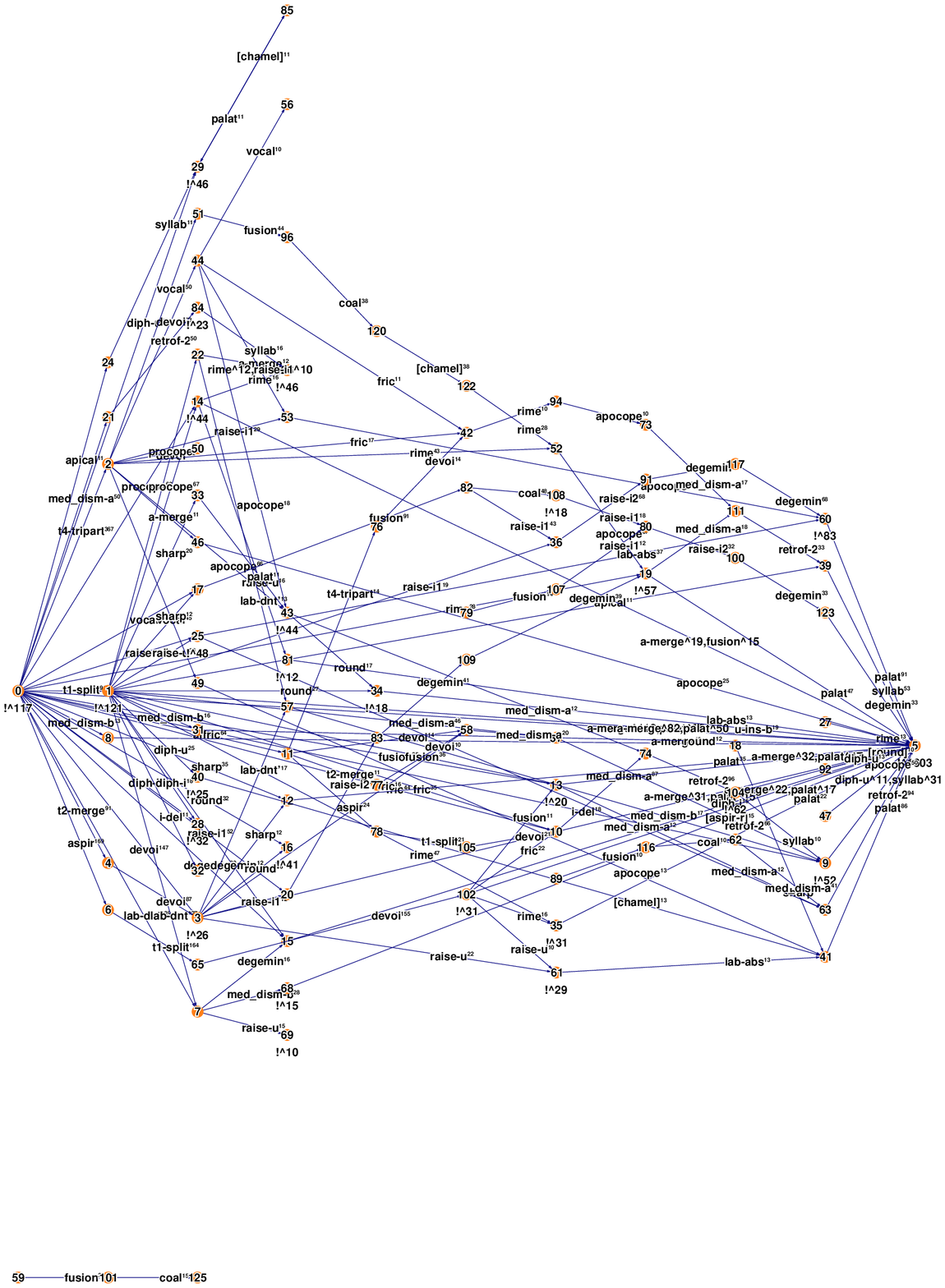,width=13cm,height=21cm,clip=}}
  \caption{Reduced PFSA for the diachronic phonology from Middle Chinese
           to Modern Beijing (Allophonic detail excluded)}
\label{fig:Mand-opfsa}
\end{figure}

\begin{figure}[htb]
  \centerline{\psfig{figure=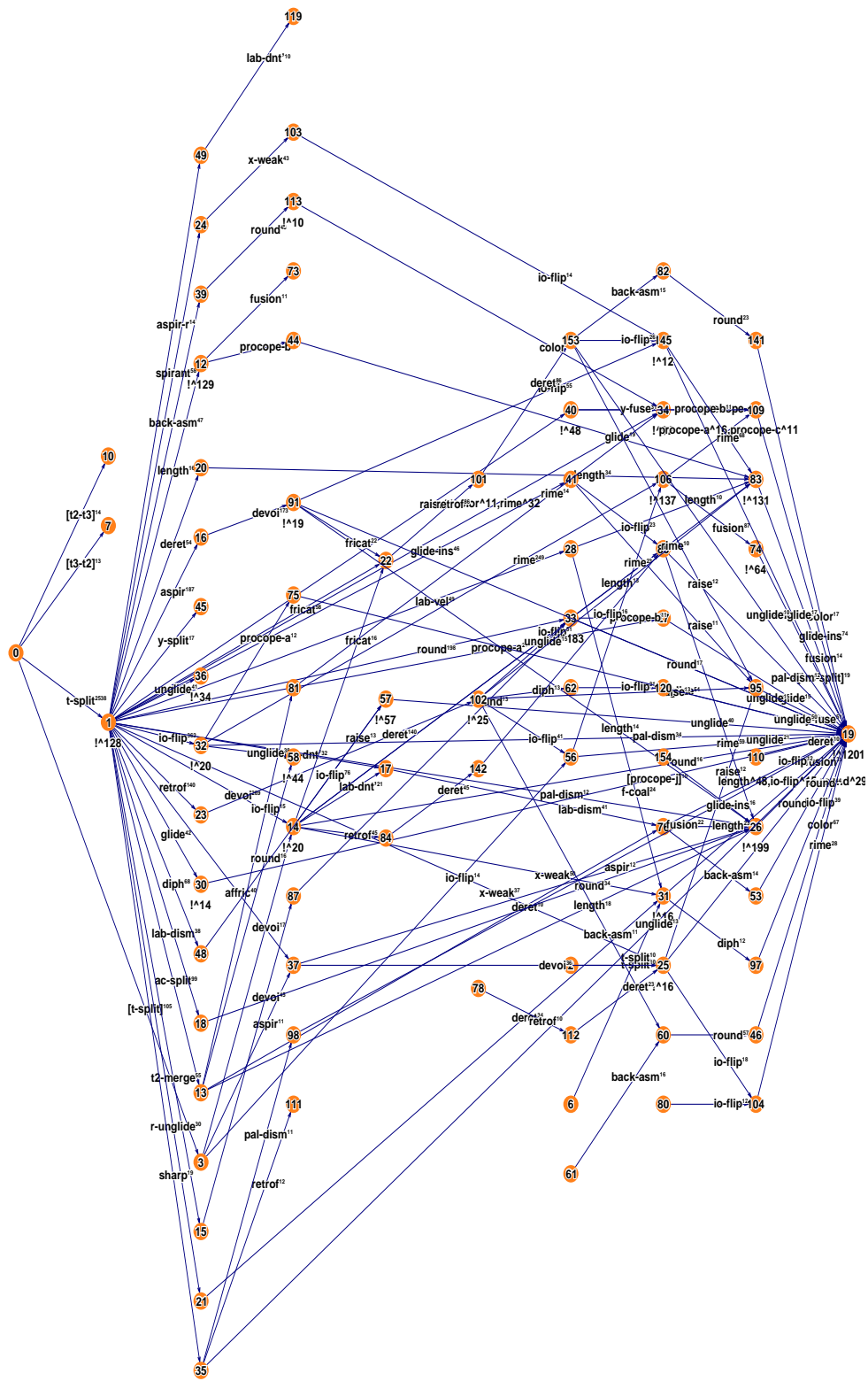,width=13cm,height=21cm,clip=}}
  \caption{Reduced PFSA for the diachronic phonology from Middle Chinese
           to Modern Cantonese (Allophonic detail excluded)}
\label{fig:Cant-opfsa}
\end{figure}
\end{onecolumn}

\end{document}